\documentclass{esannV2}
\usepackage{graphicx}
\usepackage[latin1]{inputenc}
\usepackage{amssymb,amsmath,array}
\usepackage{booktabs}

\begin{document}
\title{Modular Representations for Weak Disentanglement}

\author{Andrea Valenti and Davide Bacciu
%
\thanks{The work has been partially supported by the EU H2020 TAILOR project (n.952215).}
%
\vspace{.3cm}\\
%
University of Pisa - Department of Computer Science \\
Largo B. Pontecorvo, 3
56127 Pisa - Italy
}

\maketitle


\begin{abstract}
The recently introduced weakly disentangled representations proposed to relax some constraints of the previous definitions of disentanglement, in exchange for more flexibility. However, at the moment, weak disentanglement can only be achieved by increasing the amount of supervision as the number of factors of variations of the data increase. In this paper, we introduce modular representations for weak disentanglement, a novel method that allows to keep the amount of supervised information constant with respect the number of generative factors. The experiments shows that models using modular representations can increase their performance with respect to previous work without the need of additional supervision.
\end{abstract}

\section{Introduction}

Intuitively speaking, a \emph{disentangled representation} can be defined as a (usually low-dimensional) encoding of $z$ of  a data sample $x$, where distinct components of $z$ are responsible for encoding a specific generative factor of the data.
Despite the different attempts in the literature, coming up with a formal definition of what disentanglement actually is has proven more difficult than expected \cite{do2019theory}. Several works just assume that a disentangled representation is a representation in which a \emph{single} latent dimension responsible for encoding a \emph{single} generative factor of the data. This definition, while easy to formalise in a mathematical way, has resulted to be too restrictive in general. Recently, \cite{2022_valenti_leveraging} relaxed this definition by introducing \emph{weak disentangled} representation, where each generative factor can be encoded in a different region of the latent space without imposing additional limitations on their dimensionality. Despite the advantages of this new approach, the initial implementation of \cite{2022_valenti_leveraging} suffered from the fact that the number of annotations required for achieving weak disentanglement grew very quickly in the number of generative factors.

In this paper, we address this limitation by introducing \emph{modular representations} for weak disentanglement. In a modular representation, each partition of the latent space encodes the respective generative factor in a different adaptive prior distribution, independent from the others. We show that models that use modular representations are able to accurately perform controlled manipulations of the learned generative factors of the data without the need of increasing the amount of supervised information.

\section{Related Works}
Early methods for disentanglement are mainly concerned with increasing prior regularisation of the loss function \cite{2017_higgins_beta-vae, 2018_burgess_understanding}. 
Another line of work \cite{2018_kim_factor_vae, 2018_chen_beta-TCVAE, 2019_zhao_infovae}
penalises different terms of the same loss function in various ways. They define disentanglement using simple mathematical notions (e.g. total correlation of the latent dimensions).
After the results of
\cite{2019_locatello_challenging}, showing that pure unsupervised disentanglement is in general impossible to achieve, many works started using various degrees of supervised information 
\cite{2017_lample_fader},
either in the form of complete supervision on a small subset of training data \cite{2019_locatello_few-labels} or partial annotations on a subset of generative factors \cite{gabbay2021image}. Some works use additional classifier networks on the latent space in order to separate different parts of the latent codes. While useful, these method are not practical when multiple factors of variations need to be disentangled at the same time. Other methods for introducing implicit supervision involve dropping the i.i.d.  assumption by leveraging relational information between the samples. The relational information can be group-wise \cite{locatello2020without_compromises}, pair-wise \cite{chen2020pairwise_similarity}, or structural \cite{bai2021contrastively}.
Recently, \cite{2022_valenti_leveraging} introduced the concept of weak disentanglement, overcoming many of the above limitations. However, their method requires an increasing amount of supervision when the number of generative factors increases.

\section{Modular Representations for Weak Disentanglement}
\begin{figure}
    \centering
    \includegraphics[width=0.8\linewidth]{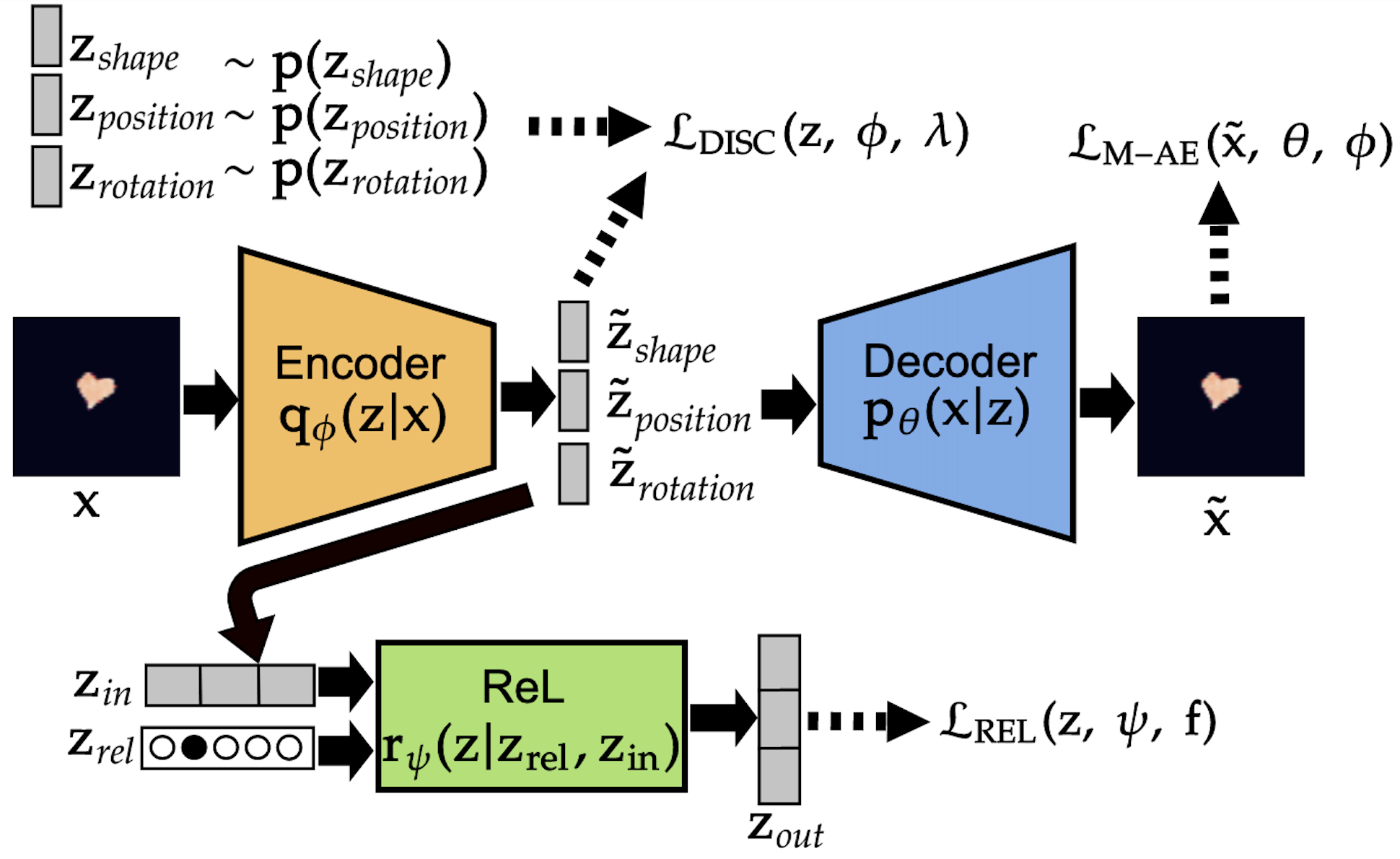}
    \caption{Overview of the model's architecture.}
    \label{fig:arch}
\end{figure}
\label{sec_modular_repr}
A general overview of the model's architecture is illustrated in Fig. \ref{fig:arch}. We frame our representation learning problem as an auto-encoding task. Given a data sample $x \sim p(x)$, we want to output a faithful reconstruction $\hat{x}$. The \emph{encoder} network $q_\phi(z|x)$, parameterised by $\phi$, takes a data sample $x$ as input and produces $G$ latent codes, where $G$ is the number of generative factors of the data. Conversely, the \emph{decoder} network $p_{\theta}(x|z_1, z_2, ..., z_G)$, parameterised by $\theta$, combines these partial latent codes to reconstruct the initial input. The resulting \emph{Modular AutoEncoder} (M-AE) model is then trained using the following maximum likelihood objective:
\begin{align}
\label{eq_aae_obj}
\max_{\theta, \phi} \mathcal{L}_{M-AE}(x, \theta, \phi)
= \mathbb{E}_{q_\phi(z|x)} \left[ \log p_\theta(x|z) \right] - \beta \sum_{g_i=1}^G{\mathrm{KL}\left(q_\phi(z_{g_i}) || p(z_{g_i}) \right)}.
\end{align}
The first term of Eq.\ref{eq_aae_obj} is directly responsible for ensuring a good reconstruction of the original input $x$. The second term, a sum of KL divergences between the aggregate posteriors $q_\phi(z_i)$ and the priors $p(z_i)$, encourages each partition of the latent space $z_i$ to follow a specific prior distribution. These priors are directly inspired from the data and are enforced in an adversarial way, similar to GANs \cite{2014_goodfellow_gan}. In particular, this term is optimised via an additional discriminator network $d_\psi(z)$, parameterised by $\lambda$:
\begin{align}
\min_\phi \max_\lambda & \quad \mathcal{L}_{DISC}(z, \phi, \lambda) = \mathbb{E}_{q_\phi(z)}[ \log d_\lambda(z)] + \mathbb{E}_{p(z)}[\log(1-d_\lambda(z))].
\end{align}
This adversarial loss allows us to choose the most suitable prior distribution for each partition. In particular, since our goal is to identify all possible real-world instances of a particular value of a generative factor $g_i$, we model each $p(z_i)$ as a \emph{mixture of normal distributions}:
$
    p(z_{g_i}) = \frac{1}{V_{g_i}}\sum_{v=1}^{V_{g_i}} \mathcal{N}(\mu_{g_i, v}, \Sigma^2_{g_i, v}),
$
where $V_{g_i}$ is the number of values that factor $g_i$ can take. We build a different mixture for each partition. Specifically, the parameters $\mu$ and $\Sigma^2$ of each prior's components are empirically estimated using a small subset of annotated samples:     $\forall i, v. \mu_{i,v} = \mathbb{E} \left[ \{z\}_{g_i=v} \right] , \quad
    \Sigma_{i, v}^2 = \mathrm{Var}\left[ \{z\}_{g_i=v} \right]$,
where $\{z\}_{g_i=v}$ denotes the subset of (encoded) supervised samples where the factor $g_i$ takes value $v$. Since each latent partition encodes a different generative factor, when can re-use the same annotated samples for computing the different parts of the prior. The second part of the model is the \emph{Relational Learner} (ReL). During training, the ReL learns how to perform controlled changes to specific properties of the data sample by leveraging the representations learned by the M-AE. 
The ReL is composed of the \emph{relational sub-network} $r_\psi(z | z_{rel}, z_{in})$, parameterised by $\psi$. Assuming that the relation to be learned affects only the value of a single factor of variations, the relational objective becomes the following:
\begin{align}
\label{eq_rel_obj}
    \max_\psi \quad \mathcal{L}_{ReL}(z, f, \psi)
    = \log p\left(z_{f(g_i)}\right) \sum_{j \ne i} \log p\left(z_{g_j}\right)
\end{align}
where $z = \left[ z_{g_1} \ldots, z_{g_i} ,\ldots z_{g_G} \right] \sim r_\psi(z | z_{rel}, z_{in})$ is the output of the relational learner. The function $f$ defines the ``connections'' between the prior components that correspond to a specific relation. This can be easily extended for losses that affect multiple factors.
This loss function encourages the partition affected by the relation to match the prior of the new value of that factor, while the other partitions remain unchanged. The correspondence between components of the prior and generative factor values is made possible by the representation learned by the M-AE. Finally, training is done end-to-end by combining the previous losses 
$
    \mathcal{L} = \mathcal{L}_{M-AE}(x, \theta, \phi, \lambda) + \mathcal{L}_{ReL}(z, \psi, f),
$

\section{Experiments}
\paragraph{Datasets.}
We consider two disentanglement tasks based on the dSprites \cite{dsprites17} and Shapes3D \cite{3dshapes18} datasets, containing respectively 2D and 3D images of shapes that express different combinations of generative factors (\texttt{shape}, \texttt{x/y-position}, \texttt{scale}, and \texttt{orientation} for dSprites; \texttt{floor-color}, \texttt{shape-color}, \texttt{background -color}, and \texttt{orientation} for Shapes3D). We consider all the relations that affect the change of a single generative factor of the data (e.g. \texttt{move-left, \texttt{move-right}, \texttt{+hue}, \texttt{change-shape}, etc.}). No restriction is imposed on the nuisance factors, that are able to vary freely when applying relations on the latent codes. For each dataset, we construct three versions of increasing complexity, characterised by different choices of relevant and nuisance factors.

\paragraph{Training Setting.}
The M-AE encoder and decoder are implemented as a CNN\footnote{All the code of the models and the experiments is publicly available: \texttt{ https://github.com/Andrea-V/Weak-Disentanglement }.}, while the prior's Discriminator and the ReL are 3-layers MLP with 1024 units each. We use 8-dimensional latent codes for each generative factors, for a maximum size of the latent space of $N_z=32$. All tasks use a batch size of 1024 for the M-AE and 128 for the ReL. The parameter $\beta$ of Eq. \ref{eq_aae_obj} is set to $0.1$.
The optimiser used for all modules is Adam with a learning rate of $10^{-4}$. Training is divided in two stages. In the first stage, called \emph{warmup}, only the M-AE is trained. The prior is set to $\forall i.p(z_i) \sim \mathrm{Uniform}(-1, 1)$. After $1000$ epochs we enter the \emph{full training} stage, where the prior of each latent partition is set to the adaptive prior described in Sec. \ref{sec_modular_repr}. We construct a different prior for each generative factor, leveraging the annotations of the supervised subset. At the same time, the training of the ReL begins: the input data samples are constructed as triples $\left( z_{in}, z_{rel}, z_{out} \right)$, where $z_{in}$ and $z_{out}$ are respectively the encoded input and output samples for the relation $z_{rel}$.
The latent codes are sampled from their respective components in the latent space.
The concurrent training of the M-AE and the ReL is carried on during the \emph{full training} phase for 5000 additional epochs.

\begin{table}
    \centering
    \caption{Relational accuracy of the ReL. $\tau$ is the number of supervised samples used during training.}
    \label{tab_rel_accuracy}    
    \begin{tabular}{lccccccc}
        \toprule
        & & \multicolumn{2}{c}{Previous Work \cite{2022_valenti_leveraging} } & \multicolumn{2}{c}{This Work} \\
        \cmidrule(lr){3-4}\cmidrule(lr){5-6}
        & Factor Combinations & Accuracy & $\tau$ & Accuracy & $\tau$ \\
        \midrule
        dSprites.v2 & 27    & 0.592 &  270   & \bf 0.724 & 1000 \\
        dSprites.v3 & 135   & 0.571 & 1350  & \bf 0.751 &  1000 \\
        dSprites.v4 & 1080  & 0.491 & 10800 & \bf 0.685 &  1000 \\\midrule
        Shapes3D.v2 & 40    & 0.275 &  400   & \bf 0.690 & 1000 \\
        Shapes3D.v3 & 120   & 0.220 & 1200  & \bf 0.653 &  1000 \\
        Shapes3D.v5 & 12000 & 0.124 & 12000 & \bf 0.633 &  1000 \\
        \bottomrule
    \end{tabular}
\end{table}

\paragraph{Latent Codes Manipulation.}
In this first set of experiments, we are interested to analyse how well suited are the modular representations to perform controlled changes of generative factors in the latent codes. We compute the relation accuracy of the ReL by first sampling a latent code  from the prior, then we apply a random relation and check the outcome. The results are reported in Table. \ref{tab_rel_accuracy}, compared with the previous work of \cite{2022_valenti_leveraging}. The results show that modular representations are beneficial for the accuracy of the ReL, while not requiring an increasing amount of supervised data when the number of factor value combinations increases.

\begin{table}
   \centering
   \caption{Disentanglement scores of latent representations. Higher is better.}
   \label{tab_disentanglement}
    \begin{tabular}{lcccccc}
        \toprule
         & \multicolumn{3}{c}{dSprites} & \multicolumn{3}{c}{Shapes3D} \\
         \cmidrule(lr){2-4}\cmidrule(lr){5-7}
         & DCI & MIG & SAP & DCI & MIG & SAP \\
        \midrule
 Locatello et al. \cite{2019_locatello_few-labels} & 0.533 & 0.01 & 0.01 & 0.48 & 0.05 & 0.08 \\
    Gabbay et al. \cite{gabbay2021image} & 0.8366 & 0.14 & 0.57 & \textbf{1.0} & 0.3 & \textbf{1.0} \\
        Valenti et al. \cite{2022_valenti_leveraging} & 0.9543 & \textbf{0.994} & 0.7728 & 0.6921 & 0.6897 & 0.5007\\
    Ours & \textbf{0.9732} & 0.9721 & \textbf{0.7877} & 0.7056 & \textbf{0.6919} & 0.5511 \\
    
        \bottomrule
    \end{tabular}
\end{table}

\paragraph{Disentanglement Scores.}
We compare the SAP \cite{2017_kumar_variational}, DCI \cite{2018_chen_beta-TCVAE} and MIG \cite{2018_eastwook_framework} disentanglement scores against several models of the literature. Following the approach of \cite{2022_valenti_leveraging}, we convert our modular representations into its corresponding generative factor values before computing the scores. This step can be done at no additional computational cost.
The results are reported in Table \ref{tab_disentanglement} showing that modular representations have a beneficial impact to all the scores, especially considering the challenging SAP score. This is a strong sign that the modular separation of weakly disentangled representations is indeed able to improve the disentanglement performance of generative models.

\section{Conclusion}
In this paper, we introduced a novel framework for learning \emph{modular weakly disentangled representations}. Modular representations encode each generative factor into a separate partition of the latent space, thus overcoming the need of requiring additional supervision when the number of value combinations of the generative factors increases.
The experiments show that modular representations allow to perform controlled manipulations to selected generative factors with high accuracy. This, in turn, results in high disentanglement scores. In the future, we wish to further enhance the expressivity of our methods by finding ways to encode continuous generative factors in a weakly disentangled way.


\begin{footnotesize}


\bibliographystyle{unsrt}
\bibliography{bibliography.bib}

\end{footnotesize}


\end{document}